%% file: main_iclr2026_conference.tex
\documentclass{article} % For LaTeX2e
\usepackage{iclr2026_conference,times}

\input{math_commands.tex}

\usepackage{hyperref}
\usepackage{url}

\title{Emergence WebVoyager: Toward Consistent and Transparent Evaluation of (Web) Agents in The Wild}

\author{Deepak Akkil \\
Emergence AI \\
\texttt{deepak@emergence.ai}
\And
Mowafak Allaham \\
Northwestern University \\
Emergence AI \\
\texttt{mowafak@u.northwestern.edu}
\And
Amal Raj \\
Emergence AI \\
\texttt{amal@emergence.ai}
\And
Tamer Abuelsaad \\
Emergence AI \\
\texttt{tea@emergence.ai}
\And
Ravi Kokku \\
Emergence AI \\
\texttt{ravi@emergence.ai}
}

\iclrfinalcopy % Uncomment for camera-ready version, but NOT for submission.
\begin{document}

\maketitle

\begin{abstract}
% Reliable evaluation of AI agents operating in complex, real-world environments requires methodologies that are robust, transparent, and contextually aligned with the tasks agents are intended to perform. This study identifies persistent shortcomings in existing evaluation practices of AI Agents that are particularly acute in web agent evaluation, as exemplified by our audit of WebVoyager, including task-framing ambiguity and operational variability that hinder meaningful and reproducible performance comparisons. To address this, we introduce Emergence WebVoyager, an enhanced version of the WebVoyager benchmark that standardizes evaluation methodology through clear guidelines for task instantiation, failure handling, annotation, and reporting. Applying this framework to evaluate OpenAI Operator reveals substantial performance variation across domains and task types, with an overall success rate of 68.6\% (much lower than the reported 87\% by OpenAI), demonstrating the utility of our approach for more rigorous and comparable web agent evaluation.

Reliable evaluation of AI agents operating in complex, real-world environments requires methodologies that are robust, transparent, and contextually aligned with the tasks agents are intended to perform. This study identifies persistent shortcomings in existing AI agent evaluation practices that are particularly acute in web agent evaluation, as exemplified by our audit of WebVoyager, including task-framing ambiguity and operational variability that hinder meaningful and reproducible performance comparisons. To address these challenges, we introduce Emergence WebVoyager\footnote{https://github.com/EmergenceAI/EmergenceWebVoyager}, an enhanced version of the WebVoyager benchmark that standardizes evaluation methodology through clear guidelines for task instantiation, failure handling, annotation, and reporting. Emergence WebVoyager achieves an inter-annotator agreement of 95.9\%, indicating improved clarity and reliability in both task formulation and evaluation. Applying this framework to evaluate OpenAI Operator reveals substantial performance variation across domains and task types, with an overall success rate of 68.6\%, substantially lower than the 87\% previously reported by OpenAI, demonstrating the utility of our approach for more rigorous and comparable web agent evaluation.
\end{abstract}

\section{Introduction}
\input{introduction}
\section{Related Work}

\input{related-work}
\section{Limitations of WebVoyager for Web-Agent Evaluation}\label{section_3}

\input{issues-with-webvoyager}
\section{Introducing Emergence WebVoyager }
\input{emergencewv}
%%% Add figure
\begin{figure}[t]
  \centering

  \begin{subfigure}{\columnwidth}
    \centering
    \includegraphics[width=0.6\linewidth]{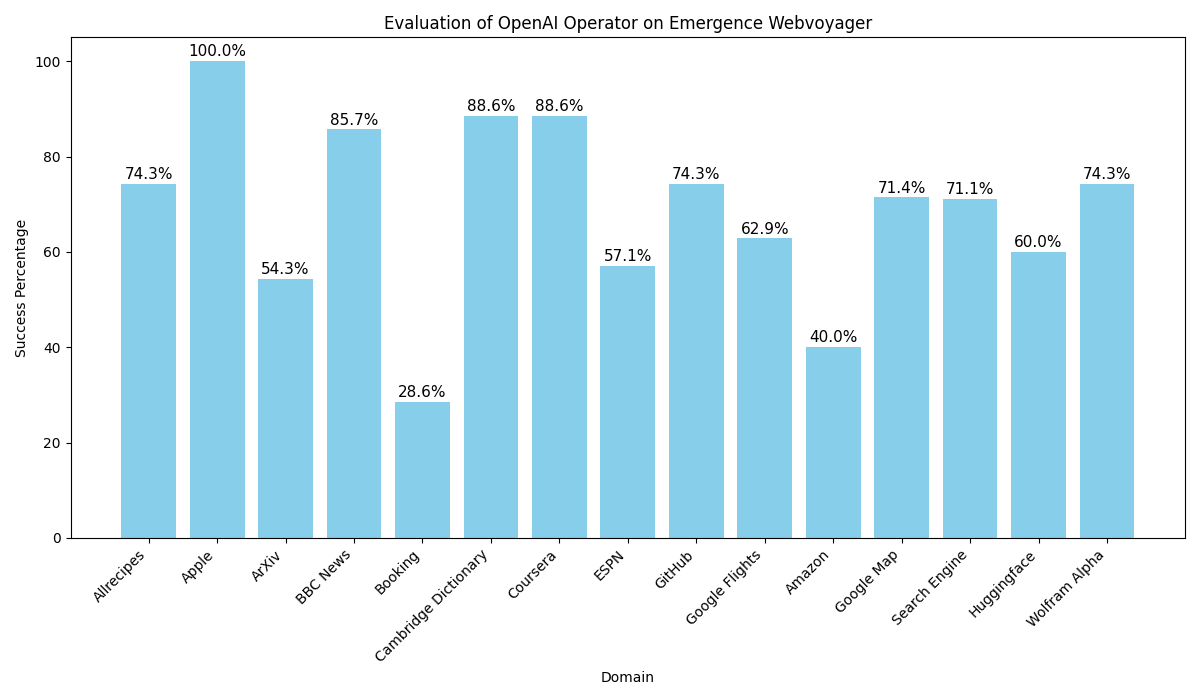}
    \caption{Success rates across domains on Emergence WebVoyager.}
    \label{fig:1a}
  \end{subfigure}

  \vspace{0.5em}

  \begin{subfigure}{\columnwidth}
    \centering
    \includegraphics[width=0.6\linewidth]{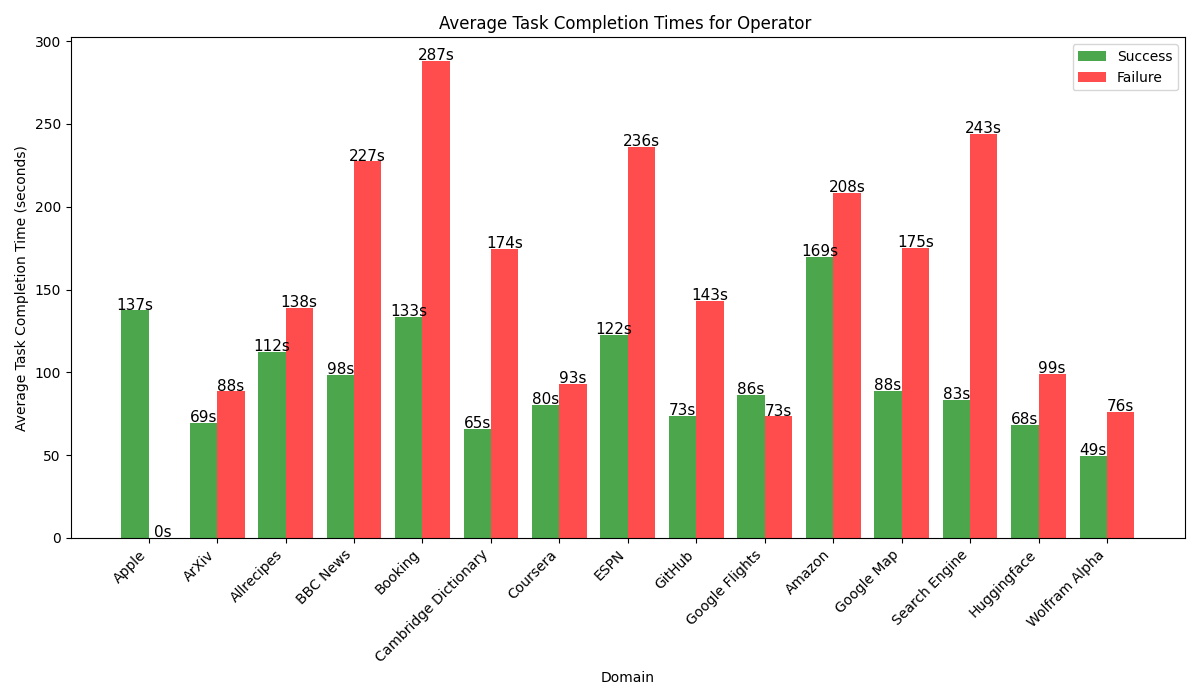}
    \caption{Average task completion time for successful and failed trajectories.}
    \label{fig:1b}
  \end{subfigure}

  \caption{Evaluation of OpenAI Operator on Emergence WebVoyager. Figure 1(a) shows the percentage of tasks successfully completed by Operator on each website. Figure 1(b) reports the average task completion times per website for successful and failed tasks}
  \label{fig:operator-eval}
\end{figure}

%%% end figure

\section{Evaluating OpenAI Operator on Emergence WebVoyager}
\input{operator-eval}
\section{Conclusion}
\input{conclusion}

\newpage
\bibliography{iclr2026_conference}
\bibliographystyle{iclr2026_conference}
\newpage
\appendix
\section{Appendix}
\input{appendix}

\end{document}

%% file: math_commands.tex
\usepackage{amsmath,amsfonts,bm}

\def\eqref#1{equation~\ref{#1}}
\def\1{\bm{1}}

\DeclareMathAlphabet{\mathsfit}{\encodingdefault}{\sfdefault}{m}{sl}
\SetMathAlphabet{\mathsfit}{bold}{\encodingdefault}{\sfdefault}{bx}{n}

%% file: introduction.tex
AI agents (i.e., AI-driven systems capable of independently performing web-related tasks) have the potential to empower humans by automating routine tasks, allowing them to devote cognitive resources to higher-level work and critical decision-making \cite{gou2025mind2web}. One category of such agents is \textit{web agents}, which complete tasks by iteratively planning and executing complex workflows that involve navigating and extracting both dynamic visual and textual information from the web. Several web agents have been developed, such as Emergence Agent-E \cite{Abuelsaad2024}, Agent-Q \citet{putta2024agent}, Runner-H \cite{hcompany2024introducingh} and Browser-Use \citet{browser_use2024}, along with other general-purpose computer control agents such as Anthropic Computer Use \cite{anthropic2024computeruse}, Project Mariner from Google \cite{mariner2025} and Operator from OpenAI. All these agents have demonstrated tremendous capabilities in performing a wide range of web tasks, as reported by numerous benchmarks.

However, the rapid improvement in the capabilities of AI agents continues to outpace existing evaluation benchmarks that rely on short or static answers \cite{deng2023mind2web}. Researchers have identified several issues with these benchmarks, including the potential exposure of large language models powering these agents to ground-truth answers due to data contamination during training \cite{deng2024unveiling,dong2024generalization,xu2024benchmark,zhang2024careful}, thereby undermining benchmark integrity \cite{han2025search}. In the context of web agents, the transient and evolving nature of information on the web paired with the open-endedness of web search and navigation may yield noisy and inconsistent performance of these agents, especially in real-world tasks \cite{deng2023mind2web}. Furthermore, existing benchmarks for evaluating the performance of such agents when operating on the web via multi-step workflows (e.g., navigating across multiple websites and pages to complete a task) exhibit persistent shortcomings in both construct (i.e., whether a benchmark measures what it claims to measure) and ecological (i.e.,  whether the task assigned to the agent truly reflects how agents are used in the wild) validity \cite{saxon2024benchmarks,liao2023rethinking,hardy2025more}. When taken together, these issues not only lead to mis-estimation of agent performance across tasks and websites, but also result in inconclusive comparisons between agents being evaluated on these tasks and websites. Such inadvertent and systemic errors in the evaluation process of these agents undermine the measurement validity, leading to inflated success metrics \cite{zhu2025establishing,koh2024visualwebarena}.

In tandem with the potential and growing adoption of AI agents for various tasks, rigorous and reproducible evaluations ensuring the reliability and trustworthiness of the outputs of these agents become indispensable. Accordingly, if existing evaluation methodologies and benchmarks of agents suffer from validity issues as well as from limitations in reliability and robustness, then the performance of these agents become open to disparate interpretations \cite{krupp2025quantifying}, potentially mischaracterizing performance gaps across various tasks, and therefore comparisons between agents become misleading. Collectively, these issues leave substantive deficiencies in agent capabilities unaddressed, hindering progress toward trustworthy and reliable autonomous web agents. 

As a step toward addressing these gaps in the evaluation of AI agents, this study presents a methodological approach for consistent and transparent evaluations, demonstrating its applicability in evaluating web agents as use case. The decision to focus on web agents is motivated by the continuously evolving and complex nature of information-seeking and retrieval on the web, which makes web agents a representative testbed for evaluating agent performance in the wild. To this end, we begin by characterizing the nature of inconsistencies in the evaluations of web-agents by manually auditing WebVoyager, a widely used benchmark that includes 643 tasks across 15 diverse websites (e.g., Amazon, GitHub, Google Flights, and Coursera). Our audit reveals 11 methodological shortcomings across the dimensions of task framing ambiguity and operational variability in execution, as elaborated on in Section \ref{section_3}. Based on these insights, we introduce Emergence WebVoyager, a  refined version of the WebVoyager benchmark that addresses these limitations and comprises 535 tasks (35 tasks per website category and 45 tasks in the search engine category). In addition, to enable and support the robust, replicable, and transparent evaluations of web agents we developed an annotation tool, a lightweight framework that is also publicly available on Github for efficient manual annotation of web agents. This framework relies on a protocol that supports modular and parametrized human-curated template of web evaluation tasks, allowing the same template to be reused across evaluations, while pairing each evaluation question with clearly defined and context-relevant success criteria for human annotators. Using EmergenceWebVoyager, we evaluate OpenAI Operator, demonstrating the viability of our annotation tool in supporting the robust assessment of agent performance on web-navigation tasks.

Our findings show that OpenAI Operator reflect substantial variation in agent performance across websites. Task success rates reach 100\% on some domains (e.g., Apple.com) while dropping as low as 35\% on others (e.g., Booking.com). Also, task completion times vary widely, ranging from 29 seconds to 1,370 seconds (approximately 24 minutes). Across the full benchmark, Operator achieves an average task success rate of 68.6\%, substantially lower than the 87\% success rate previously reported by OpenAI.

Overall, this study highlights pervasive shortcomings in the evaluation practices of web agents showcasing issues that threatens the robustness and integrity of WebVoyager \cite{he2024webvoyager}. Researchers and practitioners evaluating agents can leverage the lessons learned from our audit, the methodological approach proposed in Emergence WebVoyager, along with the accompanying annotation tool, as a baseline for designing evaluation protocols and success criteria for AI agents more broadly, enabling more robust, replicable, and transparent evaluation standards. Most importantly, such evaluations must \textit{always} be contextually aligned with the tasks agents are intended to perform. Failure to ensure this alignment can lead to misinterpretation of agent performance, as exemplified by our findings from evaluating OpenAI Operator on Emergence WebVoyager. 

%% file: related-work.tex
Evaluating the performance of web agents, like many emerging agentic systems, is non-trivial and requires rigorous methodologies that go beyond any single benchmark or fully automated evaluation protocol (e.g., LLM-as-a-Judge or Agent-as-a-Judge) to adequately assess their capabilities \cite{starace2025paperbench,zhuge2024agent,gou2025mind2web}. Growing concerns around robustness and reliability of these agents \cite{singh2025leaderboard,xue2025illusion,krupp2025web} have therefore motivated researchers to contribute and develop evaluations to better capture the diversity of tasks and complexities autonomous web agents may encounter as they complete tasks in the wild. Broadly, the methodological contributions from these efforts fall into two evaluation environments: sandboxed offline environments and live online environments, each presenting distinct advantages and trade-offs.

Sandboxed offline environments can be based on either simple mock websites (e.g., WebArena \cite{zhou2023webarena}, Visual WebArena \cite{koh2024visualwebarena}, WebShop \cite{yao2022webshop}, ST-WebAgentBench \cite{levy2024st}) or limited cached versions of real websites (e.g. Mind2Web \cite{deng2023mind2web} and across devices (e.g., OSWorld \cite{xie2024osworld}, and AndroidWorld \cite{rawles2024androidworld}). These types of evaluations offer a self-hosted containerized sites that emulate real-world environments, permitting task execution through standard browser. Although, this approach provides controlled setting that potentially supports repeatable and systematic evaluation of web agents, it falls short of representing and reflecting the full complexity of real-world web interfaces. For example, simulated websites (i.e., mocked versions) provide limited UI designs and a narrow range of interaction types, while cached versions of real sites severely restrict the agent's ability to explore freely. As a result, there are growing concerns about how well agents trained and evaluated in these environments will generalize to real-world deployments.

In contrast to sandboxed evaluation environments, live online evaluations rely on real-time interactions with the web (e.g., WebVoyager \cite{he2024webvoyager}, Mind2Web-Live \cite{pan2024webcanvas}, \cite{xue2025illusion}). These benchmarks offer a distinct advantage in evaluating web agents by capturing the complexity and diversity of real-world websites. They feature a wide range of user interface elements, such as advanced date pickers, range selectors, infinite scrolling, and modal pop-ups or banners that are ubiquitous in modern web design. More importantly, they reflect implementation details that are equally ubiquitous, including iFrames (embedded external content), Shadow DOMs (encapsulated structure and styling), and canvas elements (graphics rendered outside the traditional HTML DOM). These components represent real-world challenges that web agents must be able to navigate, yet are often missing or under-represented in simulated environments. 

Despite the potentials of evaluation benchmarks in online environments, especially those that go beyond a ``golden" answer to include ``answer" and ``trajectory" pairs for each task, they introduce a few practical challenges. First, the dynamic and constantly evolving nature of online websites makes it difficult to automatically verify whether a web agent has successfully completed a task. There may be multiple valid ways to accomplish a given goal, and expected outcomes can change frequently due to content updates. This variability introduces ambiguity in defining an agent's success and complicates the development of standardized evaluation criteria. For instance, existing automatic evaluation methods on WebVoyager suffers from 20-40\% disagreement with a human annotator \cite{xue2025illusion}. %As a result, accurate assessment often requires manual annotation, which is labor-intensive and not easily scalable. 
Second, the tasks included in the benchmarks are susceptible to becoming outdated as websites evolve or deprecate certain features, requiring ongoing maintenance to ensure that tasks remain valid and achievable. Without regular curation, the reliability and relevance of evaluation results can degrade over time. Third, comparisons between benchmark runs are confounded by variability that stems from factors such as geographic location, time of execution, and transient server-side changes, all of which can affect agent behavior in non-deterministic ways.

Focusing on WebVoyager, one of the key shortcoming of the WebVoyager benchmark is the ambiguity in the definition of success criteria across tasks, coupled with the lack of rigorous analysis of failure modes that affect evaluation quality and reproducibility. As a result, it is unclear whether reported performance of agents evaluated using this benchmark is stable or reliable across runs. For instance, previous work has reported that the evaluation accuracy can degrade significantly as task complexity increases \cite{xue2025illusion}, further underscoring the limitations of this benchmark.

Accordingly, ensuring reliable comparisons between different evaluation runs is paramount to advancing web agents, as they enable the research community to evaluate progress and identify techniques that work (and those that do not) \cite{singh2025leaderboard}. Thus, methodological rigor is critical when designing evaluations of web agents on online benchmarks to address the variability in agents' performance and ensure fair assessments between providers and across various tasks. A necessary first step toward this goal is the critical analysis of existing benchmarks.

%% file: issues-with-webvoyager.tex
By manually auditing each of the 643 tasks by two independent annotators, we conduct a critical analysis of the WebVoyager benchmark assessing the clarity of task intent and the definition of success criteria, thereby identifying two dimensions that contribute to inconsistency in web agent evaluations using this benchmark: (1) \textbf{task framing ambiguity} (i.e., tasks are often under specified or inconsistently phrased, leading to divergent interpretations by both agents and human annotators) and (2) \textbf{operational variability in execution} (i.e, evaluations have used differing environments, configurations, thus making cross-paper comparisons unreliable). In addition, we corroborate these findings by examining a range of prior studies relying on WebVoyager for evaluating web agents, focusing on their execution procedures and evaluation strategies. %We elaborate on each issue and our findings in details in subsequent sections.

\subsection{Task framing ambiguity}
Ambiguities in task definitions can lead to inconsistent interpretations by both agents and human annotators. WebVoyager showcases multiple issues associated with task framing and these fall into four main subcategories: website of execution is not enforced \ref{3.1.1}, ambiguous task definitions \ref{3.1.2}, reliance on static dates \ref{3.1.3}, and temporal sensitivity of tasks \ref{3.1.4}. Also, in section \ref{3.1.5}, we report the prevalence of tasks within WebVoyager that do not fully leverage the capabilities of a web agent, which could contribute to inflating performance metrics of various agents.
 
\subsubsection{Website of Execution is Not Enforced} \label{3.1.1} In the original WebVoyager benchmark, the website on which a task should be performed is typically implied by the \texttt{start\_url} (i.e., the URL the agent is on at the beginning of the task). While this often nudges the agent toward a specific site, it does not enforce it. If the initial attempt fails, the agent is free to fall back on alternate websites or even global search engines. Because tasks do not explicitly restrict the site of execution, agents can take many possible paths to complete them, each with varying degrees of complexity. This flexibility introduces substantial variability across different runs and agent behaviors.

However, such unconstrained execution is often unrealistic in real-world settings, particularly in enterprise environments where global search capabilities may not be available, and organizational policies or user preferences typically restrict the agent to a predefined set of approved websites and tools. One potential way to address this limitation, as specified in Emergence WebVoyager, is to explicitly specify the required website for each task. Accordingly, an agent is considered to have failed the task if it does not complete it on the designated website, thereby enforcing consistency and reflecting more realistic usage constraints.

\begin{tcolorbox}[colback=yellow!10, colframe=yellow!50!black, width=\textwidth, boxrule=0.5mm, sharp corners]
\textbf{Task from original WebVoyager:}  Find a recipe for a vegetarian lasagna that has at least a four-star rating and uses zucchini.
\vspace{1em}  % Adds vertical space between the lines

\textbf{Corresponding task from Emergence WebVoyager:} Using the website \url{https://www.allrecipes.com/}, find a recipe for vegetarian lasagna that has at least a four-star rating and uses zucchini
\end{tcolorbox}

\subsubsection{Ambiguous Task Definitions} \label{3.1.2}  Several tasks in the WebVoyager benchmark are phrased in ways that are ambigious and subjective. For example, the instruction to ``book a highly rated hotel" suggests that the agent is expected to complete a reservation, potentially involving actions like logging in or making a payment. However, the actual intent of the task is often just to find a suitable hotel. Such ambiguous language can mislead both agents and annotators, resulting in inconsistent interpretations and evaluations.
    
\begin{tcolorbox}[colback=yellow!10, colframe=yellow!50!black, width=\textwidth, boxrule=0.5mm, sharp corners]
\textbf{Task from original WebVoyager:}  Book a highly-rated hotel with a swimming pool and free WiFi near the Louvre Museum in Paris for the weekend of March 3-5, 2024.
\vspace{1em}  % Adds vertical space between the lines

\textbf{Corresponding task from Emergence WebVoyager:} Using the website \url{https://www.booking.com/}, find a highly-rated hotel with a swimming pool and free WiFi near the Louvre Museum in Paris for the weekend of [start date]-[end date].
\end{tcolorbox}

\subsubsection{Reliance on Static Dates}\label{3.1.3}  WebVoyager benchmark includes over 75 tasks that use hard-coded static dates. Over time, many of these dates have become outdated, making the tasks unachievable in their original form. A common workaround in recent evaluations has been to manually update these dates to future values. Some studies explain how they update the dates, such as adding eight months to the original date \cite{Abuelsaad2024}, while many others do not provide any details.

This lack of standardization poses a problem. Task complexity often depends on how far into the future the selected dates are, especially when interacting with calendar widgets that require agents to manually click through months. Without a consistent and transparent procedure for date selection, evaluations can vary significantly in level of difficulty, reducing the comparability and reproducibility of results.
    
\begin{tcolorbox}[colback=yellow!10, colframe=yellow!50!black, width=\textwidth, boxrule=0.5mm, sharp corners]
\textbf{Task from original WebVoyager:}  Search a hotel with free WiFi and air conditioning in Bali from Jan 1 to Jan 4, 2024.
\vspace{1em}  % Adds vertical space between the lines

\textbf{Corresponding task from Emergence WebVoyager (Before Instantiation):} Using the website \url{https://www.booking.com/}, find a hotel with free WiFi and air conditioning in Bali from \texttt[{{{@eval:(now() + timedelta(days=20)).strftime('\%B \%d \%Y')}}}] to \texttt{{{[@eval:(now() + timedelta(days=24)).strftime('\%B \%d \%Y')}}}].

\vspace{1em}  
\textbf{Corresponding task from Emergence WebVoyager (After Instantiation)}: Using the website https://www.booking.com/, find a hotel with free WiFi and air conditioning in Bali from May 20 2025 to May 24 2025

\end{tcolorbox}

\subsubsection{Temporal Sensitivity of Tasks} \label{3.1.4} Several tasks in the WebVoyager benchmark is sensitive to the time of execution and, by extension, to the time zone in which the agent or user is operating. These tasks often involve constraints like finding businesses that are currently open or events happening ``now", which can yield different results depending on when and where the task is executed.

\begin{tcolorbox}[colback=yellow!10, colframe=yellow!50!black, width=\textwidth, boxrule=0.5mm, sharp corners]
\textbf{Task from original WebVoyager:}  Search for locksmiths open now but not open 24 hours in Texas City.
\vspace{1em}  % Adds vertical space between the lines

\textbf{Corresponding task from Emergence WebVoyager:} Using the website \url{https://www.google.com/maps/}, search for locksmiths open at 10.00AM but not open 24 hours in Texas City.

\end{tcolorbox}
    
\subsubsection{Prevalence of Relatively Easy Tasks in the Original Benchmark}\label{3.1.5} 
Many tasks ($\approx$15\%) in the original WebVoyager dataset do not fully leverage the capabilities of a web agent. Their answers are often static, commonly known facts that are likely present in the pretraining data of most large language models (LLMs), or are easily retrievable via a simple search engine query. For example, tasks such as ``Find out the current world record for the men's 100m sprint" or ``Find the initial release date for Guardians of the Galaxy Vol. 3" can probably be answered by LLMs using their training data without navigating the web or interacting with complex interfaces. Accordingly, the prevalence of such easy tasks tend to saturate the benchmark, conflating agentic capabilities in web navigation with an LLM memorization or shallow retrieval of its training data. 

\begin{tcolorbox}[colback=yellow!10, colframe=yellow!50!black, width=\textwidth, boxrule=0.5mm, sharp corners]
\textbf{Removed Task from Original WebVoyager:} \\
Find out the current world record for the men's 100m sprint.

\vspace{1em}

\textbf{New Task Added in Emergence WebVoyager:} \\
Using any search engine as a starting point, find the earnings press release for Adobe for Q4 2023. What does the press release state about the Earnings Per Share (EPS) target for Q1 2024?
\end{tcolorbox}

\subsection{Operational Variability in Execution}
The WebVoyager benchmark has been executed under diverse operational conditions across different research efforts, introducing substantial variability that undermines direct comparisons of different evaluations. In this section, we detail issues related to the location of benchmark execution, as well as how constraints and agent failure modes are handled during evaluation. A summary of how these factors are handled across various agents is provided in Table~\ref{tab:ComparisonTransposedCentered}.  

\input{performance-table}
\vspace{1em}  % Adds vertical space between the lines

\subsubsection{Location of Benchmark Execution}
Many websites in the WebVoyager benchmark implement localization features that can either hide or significantly alter the accessibility of information or capabilities required to complete a task. For example, ESPN may present entirely different layouts, showing or hiding sports such as NHL from its homepage depending on user's location. This results in varying levels of task complexity for the same benchmark executed from different geographical regions. Additionally, sites like Amazon apply language localization, while others may enforce legal requirements in certain regions (e.g., mandatory cookie consent banners in the EU), introducing further variability that web agents may have not been optimized to handle. Consequently, the under-reporting of the regions from which web agents are executing tasks from will likely have implications on the interpretation of the agents' performance. Yet, many published studies using WebVoyager do not disclose such information. Thus, for benchmarking of web agents to be consistent and comparable, it's essential to control location-related variability during execution and account for such variability in performance evaluation. One approach to address this issue is to run agents from a specific country (e.g., the United States), either by using a VPN or selecting a cloud infrastructure that is present in that country.

% \begin{tcolorbox}[colback=yellow!10, colframe=yellow!50!black, width=\textwidth, boxrule=0.5mm, sharp corners]
% \textbf{Usage Guidelines:} 1. Emergence WebVoyager must be executed from within the United States, either by using a VPN or selecting a U.S.-based cloud infrastructure region.
% \end{tcolorbox}

\subsubsection{Handling of External Constraints such as CAPTCHAs}

External constraints, such as CAPTCHAs, rate-limiting, or temporary website outages, are realistic challenges that web agents may encounter in real-world settings. However, how these issues are handled during benchmarking can significantly affect evaluation outcomes. To ensure fair and reliable comparisons across agents, the impact of such external factors should be minimized.

While the WebVoyager benchmark aimed to exclude websites that trigger CAPTCHAs, this is becoming increasingly difficult to guarantee. Many commonly used sites, such as Amazon and Cambridge Dictionary, now employ bot-detection systems that intermittently present CAPTCHAs or deny access when automated behavior is suspected. As web agents become more prevalent, such restrictions are likely to increase. Importantly, expecting agents to bypass CAPTCHAs autonomously is both unrealistic and ethically problematic.

There is currently a lack of standardization in how evaluations treat these failures. For instance, systems like Wilbur and Agent-E count CAPTCHAs or server-side downtime as legitimate task failures, while other evaluations do not report such issues at all (see Table \ref{tab:ComparisonTransposedCentered}). These inconsistencies can skew reported success rates and undermine the comparability and fairness of benchmark results.

% In addition, a common method used to enhance the robustness of an evaluation on a benchmark, is for AI agents is to run multiple times on the task and report the best result among $N$ runs, acknowledging the inherent stochasticity of these systems. %However, this practice can introduce significant bias if not properly controlled. When using Emergence WebVoyager, re-runs should be explicitly restricted to cases of external failure only.

\subsubsection{Differences in Evaluation Protocols}
Evaluation protocols vary significantly across studies that use WebVoyager as a benchmark. For instance, agents such as Agent-E \cite{Abuelsaad2024} and Project Mariner \cite{mariner2025} rely on fully manual evaluation. In contrast, others like Browser Use \cite{browser_use2024} and Convergence \cite{convergence_proxy2025} adopt a hybrid approach, combining automated evaluation with selective human review. One example on such a hybrid approach is Browser Use utilization of a custom evaluator that assign a label to outcomes as either ``success", ``failure", or ``unknown". Tasks marked as ``unknown" or ``failure" were then manually reviewed and corrected. However, this improvised evaluation process inherently favors higher reported success rates, as it overlooks false positives (Type-I errors) by not reviewing tasks initially labeled as ``success". Moreover, the use of fully automated evaluation methods without human oversight introduces other set of issues that threatens the credibility of such evaluations because current automatic evaluators have not been rigorously validated in terms of repeatability or their distribution of Type-I and Type-II errors. In fact, state-of-the-art automatic evaluators for WebVoyager exhibit disagreement rates of 20–40\% compared to human annotations \cite{xue2025illusion}. Given that top-performing web agents may differ in performance by only a few percentage points, relying on such error-prone automatic evaluators hinders the progress towards fair and trustworthy evaluation of web agents. %At present, there is no reliable or widely accepted automatic evaluation protocol for WebVoyager that has been rigorously validated for repeatability, robustness to error modes, and error rates within a few percentage points.

\begin{tcolorbox}[colback=yellow!10, colframe=yellow!50!black, width=\textwidth, boxrule=0.5mm, sharp corners]
\textbf{Recommendation:} We recommend relying on manual annotation methods for evaluating web agents until a more rigorously validated automatic evaluator becomes available. To enable more robust evaluations, we are open-sourcing a simple \href{https://github.com/EmergenceAI/EmergenceWebVoyager}{\textcolor{blue}{annotation tool}} that can be extended beyond the evaluation of Emergence WebVoyager.
\end{tcolorbox}

\subsubsection{Interpretation of Task Success Can Be Subjective}

The evaluation of a task success in WebVoyager can be influenced by subjective human interpretation, introducing variability in how tasks completed by web agents are judged as successes or failures. One source of this subjectivity stems from how annotators interpret the linguistic cues embedded in task descriptions. Phrases such as ``find a high-rated recipe'' or ``identify three innovative and widely recognized open-source projects'' may appear straightforward, but are inherently open to interpretation. What qualifies as ``high-rated'' or ``widely recognized'' can vary depending on the annotator’s background, domain expertise, or expectations.

A second source of variability arises from differing views on what constitutes successful task completion. This is particularly relevant for compound or under specified tasks. For example, in a task like ``search for the latest preprints about quantum computing'', some annotators may consider the task complete once the agent initiates a search, regardless of the outcome. Others may expect the agent to locate a relevant pre-print and explicitly present it in the final response. In the absence of standardized success criteria, such subjective judgments can lead to inconsistent annotations and reduced comparability across evaluations. This highlights the need for clearer task definitions and well-specified evaluation guidelines to ensure reliability and reproducibility in benchmark results.

\subsubsection{Task Update and Exclusion Criteria}
A known challenge of using online benchmarks is that websites evolve, functionalities may change, interfaces may shift, and existing features can be deprecated. As a result, tasks that were once achievable may later become impossible to complete. Unfortunately, the WebVoyager benchmark has not been actively maintained by its creators. Consequently, different WebVoyager-based evaluations have handled this issue inconsistently, updating and excluding varying numbers of tasks based on differing, and often subjective, criteria. For example, Browser Use \cite{browser_use2024} excluded 55 tasks it labeled as ``impossible", though many of these were merely difficult rather than truly unachievable. Skyvern\footnote{\url{https://blog.skyvern.com/skyvern-2-0-state-of-the-art-web-navigation-with-85-8-on-webvoyager-eval/}}, on the other hand, explicitly modified numerous tasks making them easier for the web agent to complete. In one case, the original instruction, \textit{``Find a five-star rated chocolate chip cookie recipe that takes ..."} was changed to \textit{``Find a chocolate chip cookie recipe that has at least a 4.5-star rating and takes..."}, since finding a perfect five-star recipe is significantly more difficult and often requires traversing multiple pages of search results. Such subjective exclusion and task update practices introduce bias, reduce consistency, and undermine the reproducibility of results. Ideally, when a task becomes unachievable, it should be formally removed from the benchmark, and the success rates of all prior evaluations should be updated accordingly. This ensures that leader boards always remain valid and comparable over time.

\subsubsection{Minimum Reporting Guidelines}
Most prior work has reported only success rates as the primary evaluation metric. While useful, success alone captures only one facet of agent performance. For certain use cases, such as user-facing assistants, task completion time (TCT) can be also critical, offering insight into how efficiently an agent navigates complex web environments.

Moreover, to ensure fairness and transparency, it is essential to make available execution videos (or full step-by-step screenshots). These provide a verifiable record of the agent’s behavior, making it possible for community to compare execution runs, analyze failure modes, and compare trajectories across systems. Without these artifacts, it is difficult to audit results or understand qualitative differences between agents.

%% file: performance-table.tex
\begin{table}[t]
\centering
\small
\setlength{\tabcolsep}{4pt}
\renewcommand{\arraystretch}{1.5}
\rowcolors{2}{gray!10}{white}

\begin{tabular}{|
    >{\centering\arraybackslash}m{0.15\textwidth} |
    >{\centering\arraybackslash}m{0.15\textwidth} |
    >{\centering\arraybackslash}m{0.16\textwidth} |
    >{\centering\arraybackslash}m{0.13\textwidth} |
    >{\centering\arraybackslash}m{0.10\textwidth} |
    >{\centering\arraybackslash}m{0.11\textwidth} |
    >{\centering\arraybackslash}m{0.10\textwidth} |}
\hline
\rowcolor{gray!30}
\textbf{Model} & \textbf{Number of Tasks} & \textbf{Evaluation Method} & \textbf{Annotators per Task} & \textbf{Execution Location} & \textbf{Captcha/Rate-limit Handling} & \textbf{Reported Success Rate (\%)} \\
\hline
WebVoyager \cite{he2024webvoyager} & 643 & Manual and Automatic & 1 (3 used for 300-task subset) & unknown & unknown & 44 (text), 57 (multi) \\
\hline
Wilbur \cite{lutz2024wilbur}& 643 & Automatic (1 screenshot) & NA & unknown & considered failure & 53 \\
\hline
Agent-E \cite{Abuelsaad2024} & 643 & Manual & 1 & India & considered failure & 73.2 \\
\hline
Agent-E (auto-validation) \cite{azam2024multimodal} & 643 & Manual & 1 & India & considered failure & 81.2 \\
\hline
Runner H \footnote{\url{https://www.hcompany.ai/blog/a-research-update}}& unknown & Automatic (5 screenshots) & NA & USA & unknown & 67 \\
\hline
Browser Use \footnote{\url{https://browser-use.com/posts/sota-technical-report}} & 588 (removed 55 'impossible' tasks) & Automatic + manual fix for fails & unknown & unknown & unknown & 89 \\
\hline
Skyvern 2.0 \footnote{\url{https://blog.skyvern.com/skyvern-2-0-state-of-the-art-web-navigation-with-85-8-on-webvoyager-eval/}} & 635 (removed 8 'invalid' tasks) & Automatic & NA & unknown & unknown & 85.8 \\
\hline
Convergence \footnote{https://web.archive.org/web/20250127090314/https://convergence.ai/introducing-proxy/} & 640 (removed 3 tasks) & Automatic (15 screenshots) + full manual check & unknown & unknown & unknown & 82 \\
\hline
Project Mariner \footnote{\url{https://deepmind.google/technologies/project-mariner/}} & removed unknown 'obsolete' tasks & Manual & 3 per task (majority voting) & unknown & unknown & 83.5 \\
\hline
Project Mariner (tree search) \footnote{\url{https://deepmind.google/technologies/project-mariner/}} & removed unknown 'obsolete' tasks & Manual & 3 per task (majority voting) & unknown & unknown & 90.5 \\
\hline
Operator \footnote{\url{https://openai.com/index/computer-using-agent/}} & removed unknown tasks & Unknown & Unknown & unknown & unknown & 87 \\
\hline
\end{tabular}
\caption{Comparison of web agents evaluated on WebVoyager}
\label{tab:ComparisonTransposedCentered}
\end{table}

%% file: emergencewv.tex
To address the aforementioned limitations in WebVoyager, we introduce \href{https://github.com/EmergenceAI/EmergenceWebVoyager}{\textcolor{blue}{Emergence WebVoyager}}, a publicly accessible and refined benchmark designed to better evaluate the capabilities of web agents on real-world tasks. We manually reviewed the full task set and removed tasks that do not meaningfully require web interaction, replacing them with more challenging examples. Approximately 10\% of tasks require deeper reasoning, multi-step navigation, or interaction with time-sensitive, dynamic content. The resulting benchmark comprises 535 tasks in total (35 per category across all websites and 45 in the search engine category), yielding a more balanced and discriminative evaluation. 

EmergenceWebVoyager also reduces task framing ambiguity by explicitly specifying the required website for each task; agents are considered to have failed if tasks are completed on an incorrect site. To address temporal brittleness, tasks are parameterized using relative time placeholders (e.g., X days in the future) and instantiated at evaluation time via a provided script, ensuring consistent difficulty while maintaining reproducibility. Tasks with inherent sensitivity to execution time were further revised to remove dependencies on when or where evaluations are conducted.

As for human intervention, Emergence WebVoyager addresses this challenge through three explicit usage guidelines: (1) Human intervention is (only) allowed when presented with CAPTCHAs on a task critical website; (2) Retries are permitted (only) when failures are clearly attributable to external issues (e.g., CAPTCHAs, rate-limiting, temporary website outages); (3) No retry is allowed for genuine errors in reasoning, execution, or navigation (including system errors and bugs in the agent). By standardizing how external constraints and retries are handled, Emergence WebVoyager promotes a more accurate and fair evaluation protocol that isolates agent performance from the noise of the web environment.

To minimize subjectivity in the assessment of agent success on a task, each task is independently reviewed by two researchers and paired with explicit, human-curated annotation guidelines that define success and failure criteria, including acceptable answer variations and required interactions. The inter-annotator agreement of Emergence WebVoyager is 95.9\%, substantially exceeding prior live web benchmarks \cite{he2024webvoyager, xue2025illusion}, indicating improved clarity and reliability in both task formulation and evaluation. For more usage details on the benchmark and submission guidelines, please refer to our Github repository.

%% file: operator-eval.tex
To demonstrate how Emergence WebVoyager can be used to evaluate web agents, we conduct an evaluation of OpenAI Operator, one of the most capable web agents, by following three step process:

\textbf{Step 1: Task Instantiation --} Use the provided instantiation script to generate the dataset. This script automatically replaces placeholder dates with dynamically computed, task-relevant dates to ensure that tasks are temporally valid at the time of evaluation.

\textbf{Step 2: Task Execution -- } Run the instantiated tasks using Operator. During execution, record a video of the full interaction, log the total task completion time, and save the final answer returned by the agent. For consistency, it is recommended for evaluations to be run using a U.S.-based IP address.

\textbf{Step 3: Manual Annotation  --} The annotation web tool provided with the dataset can be optionally used to complete the evaluation. The tool is designed to streamline the manual annotation process by allowing annotators to review the agent's execution trace, outputs, and answer the questions for annotation in a simple interface. The tool can be hosted in the cloud and annotation tasks distributed to annotators across geography. The annotation output can be stored locally or to any configured Google Cloud Platform (GCP) bucket (see Figure~\ref{fig:annotation} in the appendix for a screenshot of the annotation web tool). 

Our evaluation shows that Operator's performance varied widely across domains—achieving up to 100\% task success on some websites (e.g., www.apple.com), while dropping to as low as 35\% on others (e.g., www.booking.com; see Figure \ref{fig:1a} for detailed performance across domains). In addition, task completion times also varied substantially, ranging from as little as 29 seconds to as long as 1,370 seconds (approximately 24 minutes). A notable pattern consistent with prior findings such as those reported by \cite{Abuelsaad2024} is that web agents tend to take significantly longer on tasks they ultimately fail. This is further illustrated in Figure \ref{fig:1b}, which shows the distribution of task completion times across successes and failures.

Across the full benchmark, Operator achieved an overall task success rate of 68.6\%. This moderate success rate, despite it being among the most capable systems available, underscores the challenge Emergence WebVoyager is addressing and its contribution toward robust and realistic benchmarks for evaluating web agents in the wild.

%% file: conclusion.tex
As the complexity of web content continues to increase, evaluation benchmarks and methodologies for web agents must be both reliable and robust to prevent misleading cross-agent and cross-task comparisons. To this end, we introduce Emergence WebVoyager, an enhanced version of the WebVoyager benchmark that improves the task suite and standardizes evaluation methodology for web agents. We define and implement clear guidelines for task instantiation, external failure handling, annotation procedures, and reporting requirements, promoting consistency and reproducibility. Using Emergence WebVoyager, we evaluate OpenAI Operator and observe substantial variation in performance across domains and task types, with an overall success rate of 68.6\%, demonstrating the viability of our methodological approach and its contribution to more rigorous, meaningful, and comparable evaluation of web agents. Future work may extend this framework through the addition of curated and parameterized web task templates with well-defined success criteria, further advancing transparent and replicable evaluation standards for web agents in the wild.

%% file: appendix.tex
\begin{figure}[h]
    \centering
    \includegraphics[width=1.1\linewidth]{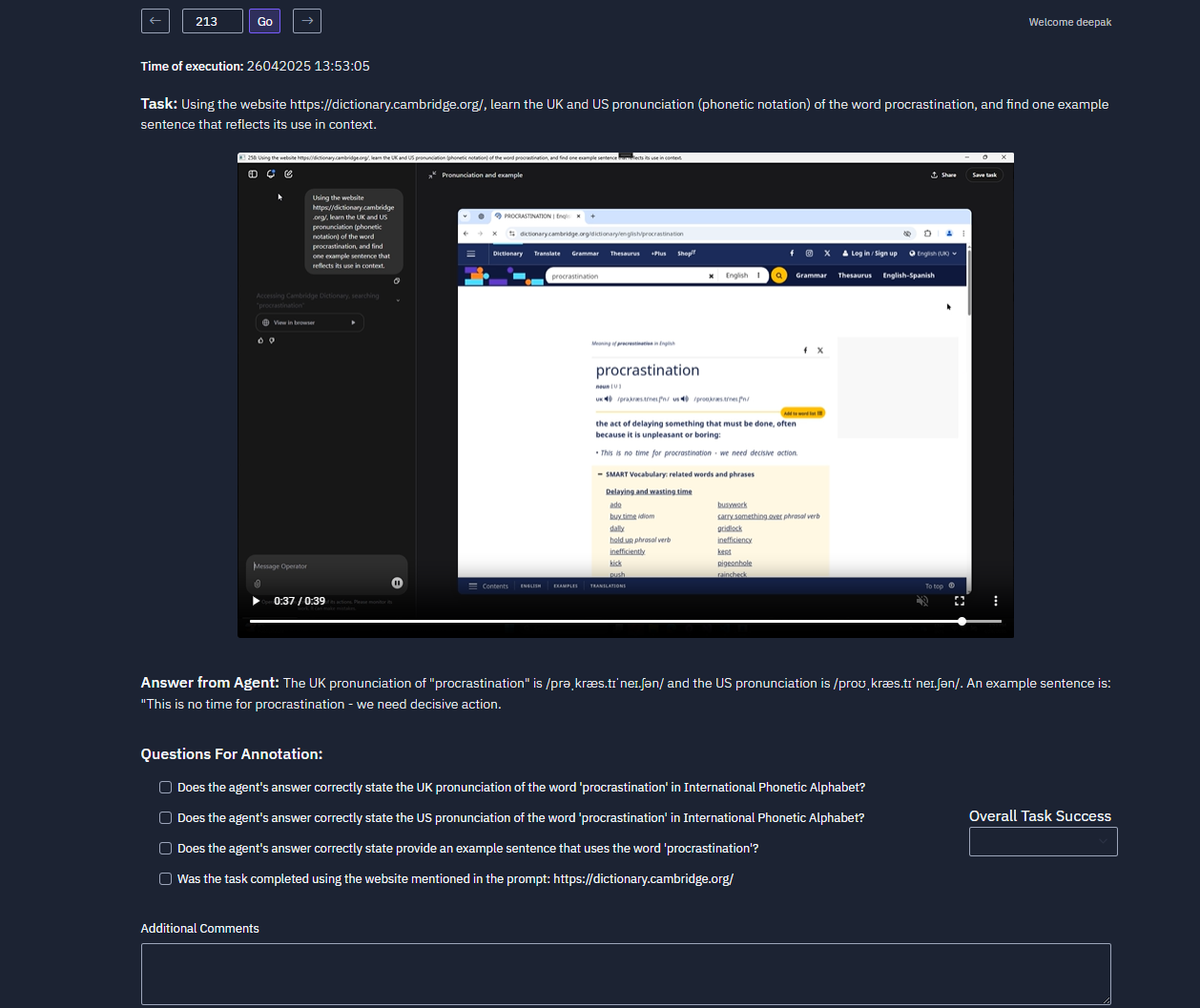}
    \caption{Annotation interface of the tool we developed and used for evaluating Operator performance in Emergence WebVoyager.}
    \label{fig:annotation}
\end{figure}